\pdfoutput=1
\documentclass[11pt]{article}

\usepackage{emnlp2021}

\usepackage{times}
\usepackage{latexsym}

\usepackage[T1]{fontenc}
\usepackage[utf8]{inputenc}

\usepackage{microtype}

\usepackage{amsmath,amssymb}
\usepackage{comment}
\usepackage{lineno}

\usepackage{xspace}
\usepackage{booktabs}
\usepackage{multicol}
\usepackage{amsmath}
\usepackage{multirow}
\usepackage{graphicx}
\usepackage{enumitem}
\usepackage{mdwlist}
\usepackage{textcomp}
\usepackage{siunitx}
\usepackage{longtable}
\usepackage[normalem]{ulem}
\usepackage[outline]{contour}

\newcommand{\jpp}[1]{\textcolor{black}{#1 }}

\newcommand{\Sref}[1]{\S\ref{#1}}

\newcommand{\fref}[1]{Figure~\ref{#1}}
\newcommand{\tref}[1]{Table~\ref{#1}}

\newcommand{\myparagraph}[1]{\noindent \textbf{#1}}
\title{Measuring Sentence-Level and Aspect-Level (Un)certainty in Science Communications}
\author{Jiaxin Pei \\
  School of Information \\
  University of Michigan \\
  \texttt{pedropei@umich.edu} \\\And
  David Jurgens \\
  School of Information \\
  University of Michigan \\
  \texttt{jurgens@umich.edu} \\
  }

\date{}

\begin{document}
\nolinenumbers

\maketitle
\begin{abstract}

Certainty and uncertainty  are fundamental to science communication. Hedges have widely been used as proxies for uncertainty. However, certainty is a complex construct, with authors expressing not only the degree but the type and aspects of uncertainty in order to give the reader a certain impression of what is known. Here, we introduce a new study of certainty that models both the level and the aspects of certainty in scientific findings.
Using a new dataset of 2167 annotated scientific findings, we demonstrate that hedges alone account for only a partial explanation of certainty. 
We show that both the overall certainty and individual aspects can be predicted with pre-trained language models, providing a more complete picture of the author's intended communication.
Downstream analyses on 431K scientific findings from news and scientific abstracts demonstrate that modeling sentence-level and aspect-level certainty is meaningful for areas like science communication. Both the model and datasets used in this paper are released at \url{https://blablablab.si.umich.edu/projects/certainty/}

\end{abstract}

\section{Introduction}

Expressing certainty about what is known is a necessary characteristic of scientific work as science involves producing knowledge about what was previously unknown \cite{friedman1999communicating, smithson2012ignorance}. Given the natural aversion to uncertainty, existing studies have found that presenting uncertainty in science communications influences people's perception of scientific findings and trust in science \cite{gustafson2019effects,Fischhoff2012CommunicatingUF,van2020effects}. Therefore, understanding how journalists and scientists communicate certainty and uncertainty is critical for understanding the current ecosystem of science journalism and further provides better guidance for uncertainty communication \cite{national2017communicating}.

\begin{figure}[t]
 \centering
  \includegraphics[width=0.48\textwidth]{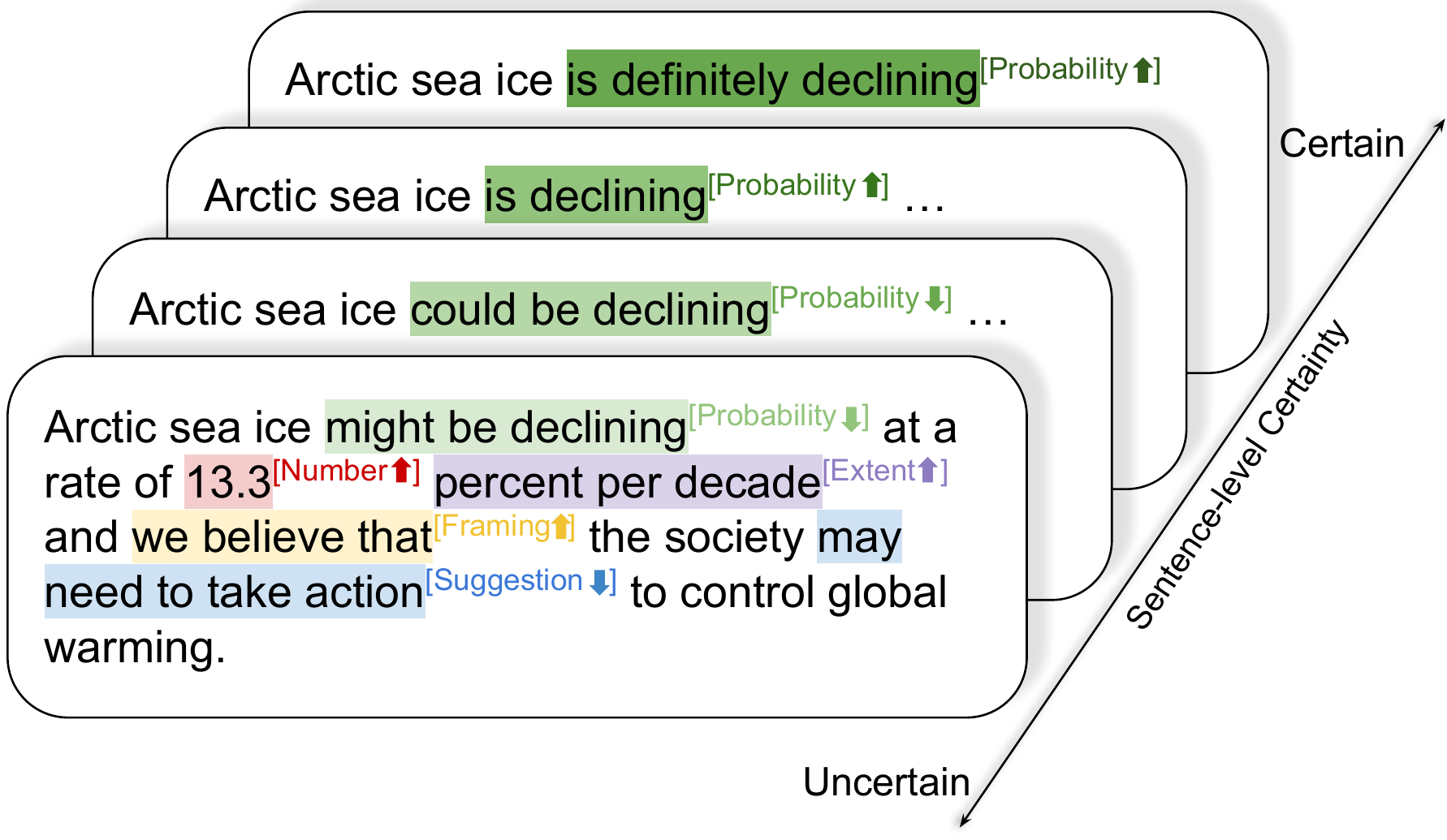}
 \caption{Certainty is a multi-dimensional construct. The certainty of a scientific finding can be perceived holistically at the sentence level from its description. However, scientific findings may involve multiple aspects that may each be described as certain (\contour{black}{$\uparrow$}) or uncertain  (\contour{black}{$\downarrow$}) (aspect-level certainty).}
\label{fig:sentence-level-example}
\end{figure}

Multiple studies in Linguistics, NLP, and the Science of Science literature have examined how certainty is expressed. These studies have modeled certainty in multiple ways, including epistemic modality \cite{vold2006epistemic}, semantic uncertainty \cite{szarvas2012cross}, verbal uncertainty \cite{hart2004verbal}, factuality \cite{sauri2009factbank}, and hedging \cite{hyland1996writing}. In practice, most uses of uncertainty rely on hedging as a coarse characterization of the overall uncertainty \cite{farkas2010conll}. However, as suggested by \citet{rubin2006certainty}, certainty itself is a complex construct and has to be modeled from multiple dimensions. The complex and subjective nature of certainty makes annotation challenging, often resulting in moderate-to-low annotator agreement \cite{henriksson2010levels,rubin2007stating}, and motivating a better and more practical way to model and annotate certainty in text. 

We propose to study certainty from the commonly-used sentence-level and, in parallel, introduce a new dimension of \textit{aspect-level 
certainty}, providing a fine-grained description of how certainty is communicated in text. This approach is analogous to work in sentiment analysis that models both holistic \cite{meena2007sentence}, and aspect-level valence \cite{schouten2015survey}---and their interactions. Based on existing categorizations of certainty, we compile six aspects of scientific findings including: \textsc{Number}, \textsc{Extent}, \textsc{Probability}, \textsc{Framing}, \textsc{Condition}, and \textsc{Suggestion}. Following carefully designed annotation guidelines and after extensive annotator training, we introduce a new annotated corpus of 2200 scientific findings equally, sampled from news and scientific abstracts for both sentence-level and aspect-level certainty, and attained reliable inter-annotator agreements. 
Analysis with this dataset suggests that the number of hedges can only partially explain the variance of the overall sentence-level of certainty (Pearson's $r$=0.55). 
Therefore, to better model certainty in scientific findings, we fine-tuned a SciBERT \cite{beltagy2019scibert} for the two tasks and it achieves 0.63 Pearson's $r$ for sentence-level certainty and an  average of 0.66 binary-F1 for aspect-level certainty. 

Our paper offers the following three contributions.
First, we provide the first dataset of scientific findings annotated with both sentence-level and aspect-level certainty and fine-tune neural language models to predict certainty in scientific findings. 
Second, using our best-performing model, we infer the sentence-level and aspect-level certainty for 431K scientific findings in news and abstracts and show that the sentence-level certainty of findings in abstracts is associated with journal impact factor and team size. Regression analysis reveals that low-impact journals and large teams often present scientific findings with higher sentence-level certainty.
Third, using 6586 findings from abstracts paired with their description in news, we find that news reports a finding with lower certainty than its corresponding description in the abstract. Fine-grained regression analysis over aspect-level certainty further reveals that  journalists describe some key aspects with less certainty. %, including \textsc{Probability}, \textsc{Number} and \textsc{Extent}. 
Despite some studies suggest that news reports tend to  describe uncertain findings as more certain  \citep[e.g.,][]{weiss1988reporting,fahnestock2009rhetoric},  our study indicates that journalists may  actually play down the certainty when  reporting  scientific findings.

\section{Modeling Certainty} 

Communicating certainty involves qualifying parts of a statement to indicate what is known and what the speaker's beliefs are. Multiple approaches have proposed models for this property of language.
The majority of these studies have focused on the degree or level of certainty \cite{rubin2006certainty}. Within this approach, hedges have been considered effective proxies for uncertainty \cite{farkas2010conll} across multiple fields including linguistics \cite{recasens2013linguistic}, economics \cite{ahir2018world}, and psychology \cite{tausczik2010psychological}, where the frequent use of hedges indicates more uncertainty and  absence indicates certainty.
Other work has proposed examining which aspects of language contribute to the perception of uncertainty. For example, \citet{rubin2006certainty} proposed a four dimension model of certainty, including perspective, focus, timeline, and level, and each dimension contains several sub-categories. 
Following, we synthesize multiple approaches for measuring sentence-level and aspect-level certainty and proposed a representative categorization.

\myparagraph{Sentence-level Certainty}
Prior research has assumed that  the level of certainty for a finding is presented, perceived, and further analyzed within one or several sentences \cite{holmes1982expressing,henriksson2010levels,rubin2007stating}. This aggregate perception represents a unified perception of various information expressed in the given piece of text and is the primary judgment of certainty along a continuum from uncertain to certain \cite{rubin2006certainty}. This perception of this overall level of certainty is known to influence people's following actions in many contexts \cite{corley2014dis,wood2009advantages}. Therefore, in modeling certainty, we include a sentence-level estimate of the certainty for a scientific finding's description.

\myparagraph{Aspect-level Certainty} 
In describing a finding, authors may contextualize which aspects are more or less certain. 
Many categorizations have been proposed to model the complex nature of these descriptions of certainty and uncertainty including types of uncertainty \cite{szarvas2012cross, thunnissen2003uncertainty}, sources of uncertainty \cite{politi2007communicating}, issues of uncertainty \cite{han2011varieties}, forms of uncertainty \cite{french1995uncertainty}, focus and perspective of certainty \cite{rubin2007stating}. While prior work focuses on various aspects of certainty, their categorizations are often domain-specific \cite{politi2007communicating,han2011varieties} or overly limited in scope to cover common aspects appearing in scientific findings \cite{szarvas2012cross,rubin2007stating}.
Here, we synthesize prior approaches and propose six representative aspects of scientific findings which could involve certainty or uncertainty, with a goal of creating a comprehensive scheme that captures most of what is seen in knowledge-intensive corpora. 

\textsc{Number} refers to certainty towards specific quantities. For example, ``approximately 250 individuals participated in this study'' is uncertain towards \textsc{Number}. Numerical information is vitally important in science communication as it is found to be the best way to promote scientific understanding in situations like climate change \cite{budescu2009improving} and health \cite{peters2014numbers}. Accordingly, the imprecision of numbers or the inaccuracy of calculations are usually considered as a form of uncertainty \cite{french1995uncertainty}. 
How to effectively communicate numerical information in scientific findings has been identified as one of the major challenges of science communication \cite{peters2014numeracy}. Identifying certainty regarding \textsc{Number} in scientific findings could help to understand how journalists and scientists communicate certainty about numbers and inspire better ways to communicate this information.

\textsc{Extent} refers to certainty about the proportion/ratio of properties that make up an object/event or the extent of a change. For example, ``This bridge is mainly composed of agate'' and ``We observe a moderate increase of suicide in Winter'' involves uncertainty towards \textsc{Extent}. %
\textsc{Extent} can be described with numbers in certain situations. For example, ``The average sea level across the world increased by approximately 30\%'' expresses extent via a number. However, unlike numbers that focus on specific quantities, \textsc{Extent} focuses on the components of an object, substance, or the extent of a change/effect. Previously, \textsc{Extent} was not explicitly proposed as a source of uncertainty, although some studies have brought up similar ideas. For example, \newcite{french1995uncertainty} considers ``Uncertainty about how much [the] impacts matter'' as a form of uncertainty and \newcite{phillips2009disclose} propose uncertainty about the strength or validity of evidence about risks, which may not be described with specific quantities. Existing studies suggest that journalists may misreport the extent of scientific findings to which it is supported by evidence \cite{dixon2013heightening}, motivating its inclusion here. 

\textsc{Probability} refers to certainty about the probability that something will occur, has occurred, or is associated with another factor. For example, "This medicine could possibly cure cancer'' and ``A is possibly associated with B'' involves uncertainty about \textsc{Probability}. \textsc{Probability} has been widely recognized as one major source of uncertainty \cite{howard1988uncertainty,mosleh1996uncertainty,politi2007communicating} and how to communicate probabilities effectively has long been an important question in science communication \cite{budescu2012effective,sinayev2015presenting}. 

\textsc{Condition} refers to the situation where something depends on a specific condition, and the condition involves certainty or uncertainty \cite{szarvas2012cross}. Scientific findings are often qualified by specific conditions under which the result is valid, which may themselves be certain or uncertain \cite{friedman1999communicating}. For example, ``Cancer could be cured if the medicine can be made shelf-stable'' is uncertain regarding \textsc{Condition}. 
    
\textsc{Framing} refers to the certainty about how scientists or journalists themselves frame or interpret the scientific finding. For example, ``We suspect A has effects on B'' involves the uncertainty from the authors while ``We conclude that A has effects on B'' frames the finding with conviction. 
This aspect is related to expressions about epistemic uncertainty\citep[the speaker having or lacking knowledge][]{szarvas2012cross,fox2011distinguishing} and psychological uncertainty in the Psychology literature \cite{windschitl1996measuring}.
In the news, journalists actively add their interpretations about the original information \cite{lin2006side} and different framing may further affect people's perception of the overall certainty of the presented information \cite{soni2014modeling}. Therefore, identifying the certainty about \textsc{Framing} could help us to better understand how journalist's framing affects people's perceptions of scientific findings.
   
\textsc{Suggestion} refers to certainty or uncertainty about the implications or future actions for the public or science community. Scientific findings do not only describe facts, but can also communicate practical implications for people's daily life \cite{batteux2021negative}. For example, ``Patients probably need more medicine to cure this disease'' involves uncertainty regarding future actions.  Uncertainty about \textsc{Suggestion} was previously identified as dynamic uncertainty in \citet{szarvas2012cross}.

A single scientific finding may include multiple aspects with their own certainties. For example, ``The vaccine is effective for 76\% of chances'' is uncertain regarding \textsc{Probability} while is certain about the specific \textsc{Number}. Similarly, ``The scientists need to do more research to understand the effect of A on B'' indicates the uncertainty about \textsc{Probability} but is certain about \textsc{Suggestion}.

\section{Data}

To study certainty, we construct a dataset of scientific findings reported in news and research articles.
News data comes from Altmetrics, which tracks mentions of scientific articles in news outlets. We restrict our analysis to U.S.-based outlets where we could retrieve the full text of the article and where the DOI for the scientific article was recorded in the Microsoft Academic Graph (MAG), which provides metadata on the article (e.g., authors, abstract, and publication venue). Supplemental material \Sref{app:data-preprocessing} contains additional details on preprocessing steps. A total of 128,942 news/article pairs were collected, spanning 273 different news outlets and 57,807 different scientific articles.

For scientific articles, we extract the findings from the abstract reported in the MAG using the abstract parser developed by \citet{Prabhakaran2016PredictingTR}, which labels sentences as background, method, introduction, result, and conclusion. We use sentences labeled as result or conclusion in our analysis. 
For news, we adopt a heuristic approach and identify all sentences containing a discovery-related keyword (e.g., find, conclude). We retain the subjective clause after the verb as the finding. Examples of findings produced by each method, as well as additional details, are reported in Supplemental Material \Sref{app:extracting-findings}. 
This process finally leads to 608,694 unique scientific findings from abstracts and 106,612 unique scientific findings from news reports. Among the 128,942 news-paper pairs, 52,406 have both identified findings from news and paper abstracts.

\section{Annotating Certainty}

We annotate scientific findings for both their sentence-level certainty and the presence and certainty of each of the six aspects. 
Given subjectivity in perception, the same expression of certainty may evoke different perceptions \cite{druzdzel1989verbal}. Prior work in annotating uncertainty has generally reported low to moderate inter-annotator agreement \cite{henriksson2010levels,rubin2007stating}; for example, \citet{rubin2007stating} reports Cohen’s $\kappa$=0.41 when annotating news certainty with a five-level Likert scale. 
To mitigate these challenges, we carefully designed the annotation procedures which are described in this section.

\myparagraph{Annotation Setup}
Annotators were recruited from a US university and received initial one-hour training. All annotators are fluent in English and have extensive experience in reading scientific news and research articles. Annotators who attained high IAA with our gold standard were retained and then went through four additional rounds of pilot annotation and discussions (2 rounds for sentence-level and 2 for aspect-level) to build consensus. All annotators were paid \$15/hr for training and annotation.

Annotation was performed in three phases. In the \textbf{first phase}, the initial data was sampled in a way to be more balanced across levels of certainty. Markers of certainty are not equally distributed throughout scientific communication \cite{rubin2006certainty}; for example, 87\% of the data labeled in \citet{henriksson2010levels} were found to be very certain.
Therefore, in the initial data, we sample 1000 findings equally from news and paper abstracts where 50\% scientific findings containing no hedges, 35\% with one hedge, and 15\% with 2+ hedges. The hedge words are collected from \citet{hyland2005metadiscourse}.

The annotators were first asked to rate how certain they perceived the finding on a six-point Likert scale as the sentence-level certainty.  Aspect-based ratings were performed in a separate round so as not to potentially bias annotators towards focusing their sentence-level judgments on the basis of aspects. For each aspect, annotators were asked to assess whether that aspect was present and, if so, whether the language for the aspect was certain or uncertain. For instances that are clearly not scientific findings, the annotators were instructed to label it as \textsc{Bad-text}. Each finding was rated by at least two annotators for sentence-level and, due to increased variance observed during training, three annotators for aspect-level. In this phase, 2349 sentence-level and 3209 aspect-level annotations were collected for 1000 findings. Annotators had a high agreement in both sentence-level and aspect-level tasks. For sentence-level, annotators attained a Krippendorff's $\alpha$=0.67, which is substantially higher than IAA for the closest comparable task \citep[][Cohen's $\kappa$=0.41]{rubin2007stating}.\footnote{While Cohen's $\kappa$ and Krippendorff's $\alpha$ are not directly comparable, both are chance-corrected scores in [-1,1] and we still view rough comparisons of magnitude useful to provide insight into annotators' relative agreement.
} 
The final sentence-level rating is computed as the average across all annotators' scores. For aspect-level certainty, the average Krippendorff's $\alpha$ is 0.57 for the six aspects, indicating moderate to high agreement.  Supplemental material \Sref{app:annotation_agreemment} contains more details about agreement scores.  We take the majority label as the final label for aspect-level certainty.  

In the \textbf{second phase}, given the label imbalance in both sentence-level and aspect-level certainty, we sample additional data for annotation based on model predictions using different sampling strategies for sentence-level and aspect-level certainties. For sentence-level certainty, we fine-tuned a roberta-base classification model and predicted all the extracted findings with sentence-level certainty. We further sampled 400 findings with low model confidence equally from news and abstracts for the second-phase annotation. For aspect-level certainty, we fine-tuned a SciBERT classification model and up-sampled 600 findings for less frequent aspects, including \textsc{Suggestion}, \textsc{Extent}, \textsc{Condition}, and \textsc{Framing}. 

Given that the first and second phase data do not reflect the distribution of certainty in a natural sample, we further randomly sampled 200 findings from all the extracted findings as to the \textbf{third phase}. Given that the annotators were capable of doing reliable annotations after the first phase, annotators independently annotated 1200 findings for the second and third phases.

\begin{figure}[t]
\centering
\includegraphics[width=2.7in]{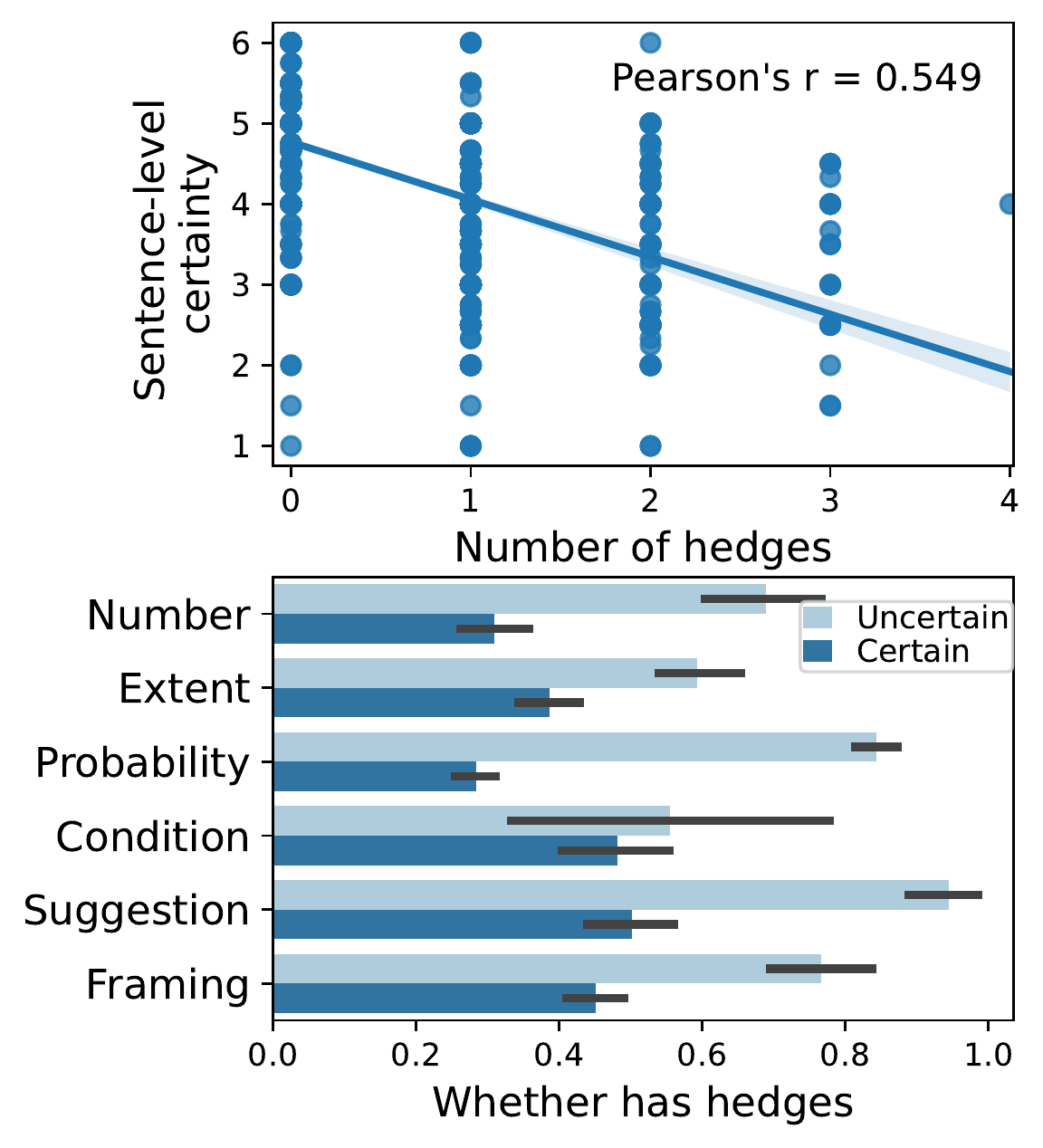}
 \caption{(Top) The number of hedges shows a moderate correlation with sentence-level certainty, while (Bottom) hedges  vary in frequency across aspects.}
\label{fig:hedge-analysis}
\end{figure}

\paragraph{Results}
The final annotated dataset contains 6958 labels for 2200 findings. After removing findings that are labeled as \textsc{Bad-text}, we obtained 1551 findings labeled with sentence-level certainty and 1760 findings labeled with aspect-level certainty, among which 1144 findings are labeled with both sentence-level and aspect-level certainty. Supplemental \Sref{app:annotated_data} present examples and distribution of the data. Previously, \citet{szarvas2012cross} considers \textsc{Condition} as one type of uncertainty, however, in the annotated data, less than 10\% of \textsc{Condition} is labeled as uncertain. This difference indicates that previously proposed types of uncertainty may not be considered as uncertain in knowledge-intensive corpus like scientific findings and demonstrates the value of aspect-level certainty.

\paragraph{To what degree do hedges capture certainty?}
Comparing the sentence-level certainty with the number of hedges (\fref{fig:hedge-analysis}, top) shows only a \textit{moderate} correlation between hedging and certainty, $r$=0.55, despite their widespread use as a proxy. For example, ``Further research is necessary to understand whether this is a causal relationship'' contains zero hedges but explicitly expresses strong uncertainty towards the causal relationship, suggesting that many descriptions of certainty are not well capture by simple hedge-based lexicons. Further, authors vary in how frequently they employ hedges when describing the different aspects of certainty (\fref{fig:hedge-analysis}, bottom). This variance in their distribution suggests that hedges are less effective as proxies for capturing uncertainty for all aspects.

\begin{figure}[t]
\centering
\includegraphics[width=2.7in]{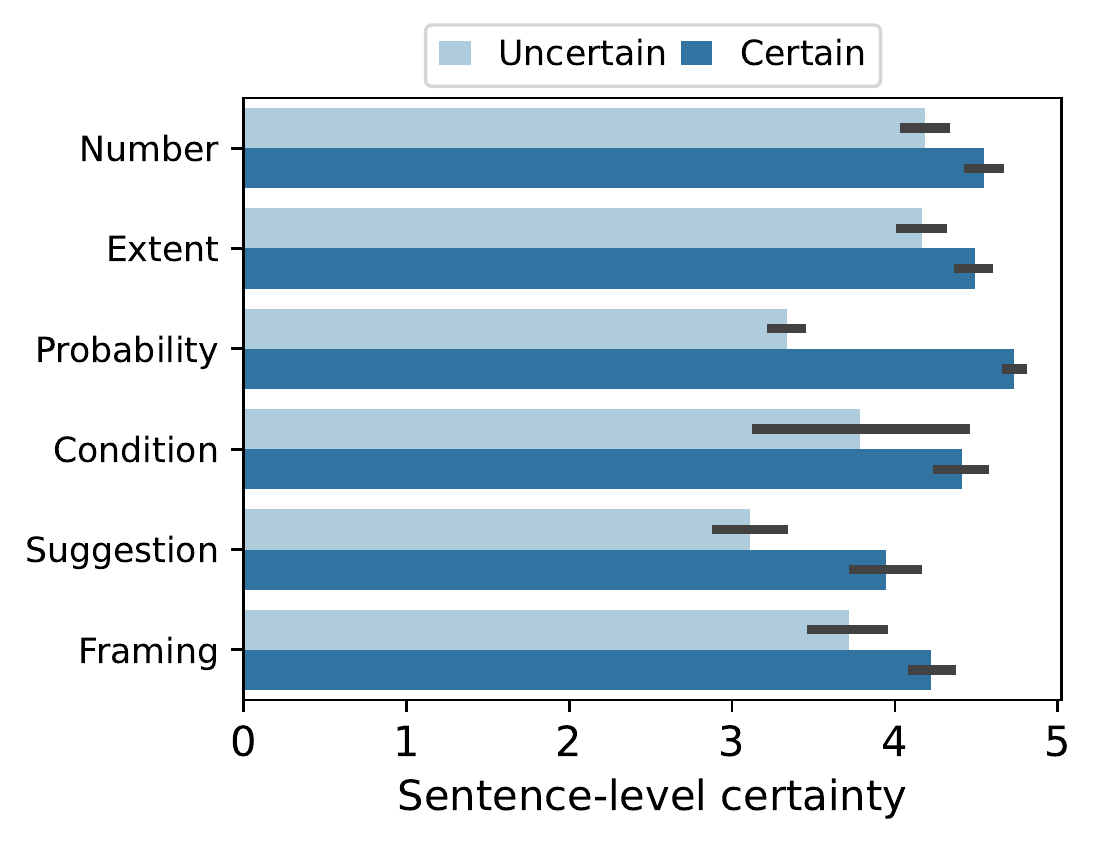}
 \caption{Relative sentence-level certainty when each aspect is certain/uncertain. The overall certainty of scientific findings is majorly affected by \textsc{Probability} and \textsc{Suggestion}, while are less affected by other aspects like \textsc{Number} and \textsc{Extent}}
\label{fig:aspects-sentence-certainty}
\end{figure}

\paragraph{What is the association between aspect-level and sentence-level certainty?}
Figure \ref{fig:aspects-sentence-certainty} shows the relative sentence-level certainty for each aspect. Uncertainties about \textsc{Probability} and \textsc{Suggestion} are associated with a sharp decrease of sentence-level certainty. However, the uncertainties about \textsc{Number} and \textsc{Extent} are only associated with a small decrease of sentence-level certainty. This result indicates that the descriptions of aspects vary in how they contribute to the perception of the overall certainty of scientific findings.

\section{Predicting Certainty}

In this section, we build models  to predict sentence- and aspect-level certainty in scientific findings to support downstream analyses of certainty. %
We test two linear baseline models and two deep-learning models based on neural language models.  As linear baselines, we include a model using bag-of-words (BoW) features and another based on the frequency of each hedging word. For neural models, we use SciBERT \cite{beltagy2019scibert}, and RoBERTa model \cite{liu2019roberta} as the base models and fine-tune them over our annotated dataset. For both sentence-level and aspect-level certainty, the data labeled in phase 1 and phase 2 are split 8:1:1 into training, validation, and test. To better reflect the expected performance generalization, the test is made from the random set annotated in Phase 3 and the 10\% test partition of Phases 1 and 2. For all the models, we also report their performance on the random test set to demonstrate their performance over the natural samples.  Supplemental Section \Sref{sec:model_details} describes additional training  details.

\begin{table}[t!]
\newcommand{\tabincell}[2]{\begin{tabular}{@{}#1@{}}#2\end{tabular}}
\centering
\resizebox{0.49\textwidth}{!}{
\begin{tabular}{r ll }

\textbf{Model} &  \textbf{$r$ on full test set} &  \textbf{$r$ on random set} \\ 
\hline
LR + Hedges        & $-$0.02 & $-$0.02 \\
LR + BoW            & \ ~~0.55 & \ ~~0.44 \\
RoBERTa-base         & \ ~~0.62 \textpm 0.051 & \ ~~0.55 \textpm 0.067 \\
SciBERT     & \textbf{\ ~~0.63 \textpm 0.061} & \textbf{\ ~~0.57 \textpm 0.065} \\
\end{tabular}
}
\caption{Sentence-level certainty  performance}
\label{tab:model_perf_level-of-uncertainty}
\end{table}

\paragraph{Sentence-level Certainty} 
We formulate sentence-level certainty prediction as a regression task for all the models and
Table \ref{tab:model_perf_level-of-uncertainty} shows the model performance. We find that a linear weighting of the hedges is unable to predict the overall sentence-level certainty when tested on the random sample, largely due to the relatively low ratio of findings containing hedges. In comparison, linear regression with bag-of-words features is better able to capture overall certainty with Pearson's $r$=0.55, suggesting that other cues in addition to hedges also affect the overall certainty in the natural sample.
Compared with the two baselines, the two neural models based on pre-trained language models achieve better performance. Both neural models are run five times with different random seeds, showing that the performance improvements over the baselines are statistically significant  (p$<$0.05 paired $t$-test).
The SciBERT model performs slightly better than the RoBERTa-base model, indicating  domain-specific pre-training is helpful though not to the point of significance. 
We use the best-performing SciBERT model ($r$=0.70) as the regressor for sentence-level certainty in the following analyses.

\begin{figure}[t]

 \includegraphics[width=3.1in]{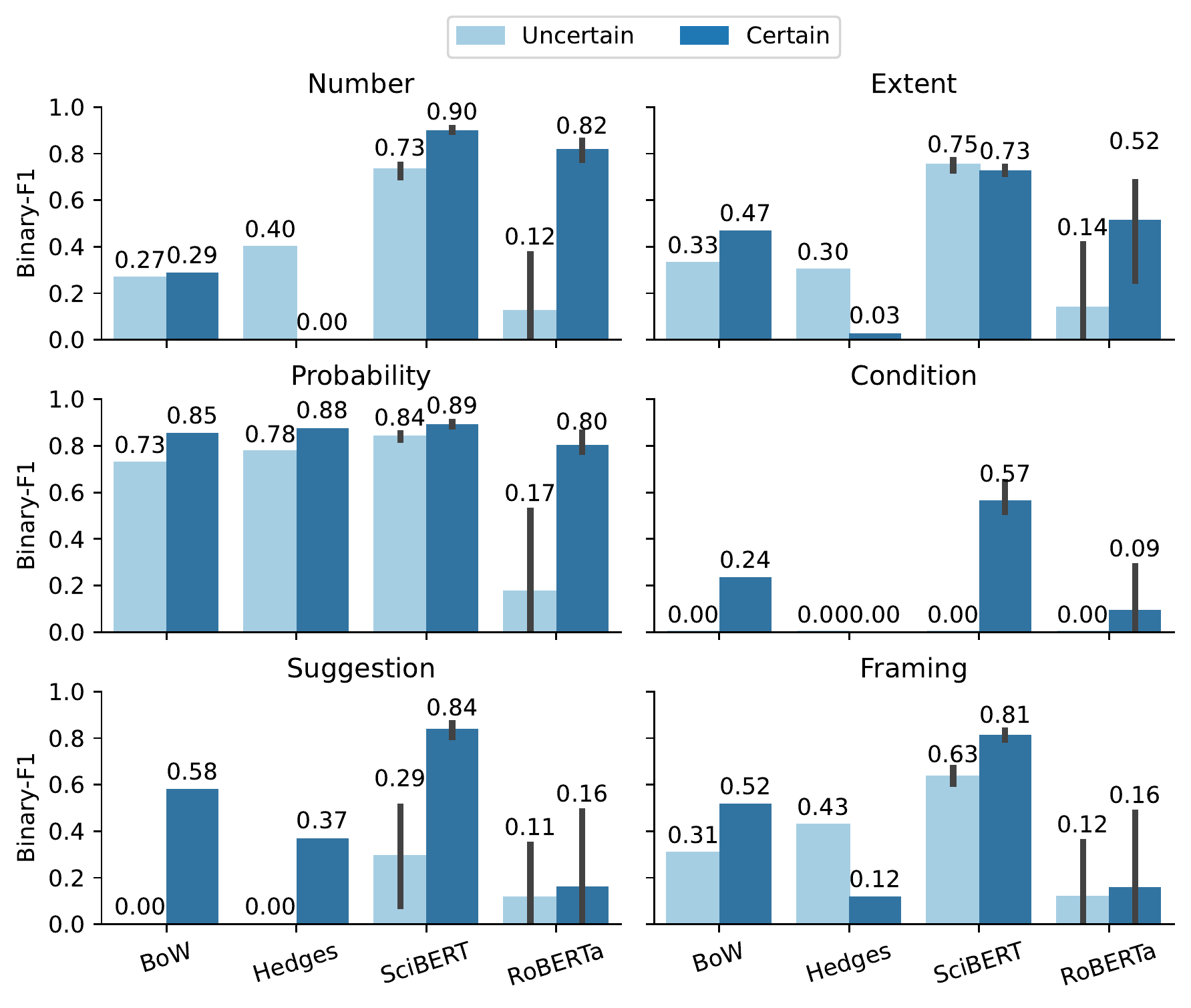}
  \caption{Binary-F1 for predicting aspect-level certainty. SciBERT outperforms all the other baselines.}
  \label{fig:model_perf_type-of-uncertainty}
\end{figure}

\myparagraph{Aspect-level Certainty} For each aspect-level certainty, we predict whether a scientific finding is \emph{NotPresent}, \emph{Certain} or \emph{Uncertain}. For the two neural models, we use shared pre-trained language models but independent classification heads for each aspect. Figure \ref{fig:model_perf_type-of-uncertainty} shows the binary-F1 scores for predicting aspect-level certainty. 
The SciBERT model consistently outperforms other baselines over the six aspects, indicating that aspect-level certainty prediction requires more domain-specific and context information. However, given that uncertainties about \textsc{Condition} and \textsc{Suggestion} are relatively rare in the annotated dataset, the SciBERT model does not capture well the uncertainty about \textsc{Condition} and shows high variance when predicting uncertainty about \textsc{Suggestion}. 
In the following analysis, we use the best-performing SciBERT model (mean F1=0.71) as the classifier for aspect-level certainty.

\section{Certainty in Science Communications}
Certainty is a core aspect of science communication \cite{friedman1999communicating} and presenting certainty in different forms (i.e., aspects) may further affect people's perception and future action about a series of issues including climate change \cite{fortner2000public} and the Covid-19 vaccine \cite{batteux2021negative}. Our models and dataset enable us to study how journalists and scientists present certainty in science communications. Here, we focus on the following five research questions. \textbf{RQ1:} Are findings in science news more certain than those in paper abstracts?
\textbf{RQ2:} Do journalists and scientists differ in their use of aspect-level certainty? 
\textbf{RQ3}: Does aspect-level certainty in abstract findings affect the sentence-level certainty in news findings?
\textbf{RQ4}: Does journal impact factor affect the certainty of scientific findings and how they are covered in news reports?
\textbf{RQ5}: Does team size affect the certainty of scientific findings and how they are covered in news reports?

RQ1--RQ3 focus on changes to the description of certainty  in the science communication process. While studies have found that news reports tend to describe uncertain findings as more certain \cite{weiss1988reporting,fahnestock2009rhetoric}, some studies suggest that news articles may also add more uncertainty to science finding in some cases \cite{friedman1999communicating}. Our model and dataset allow us to study (1) if certainty is changed, (2) if so, what aspects are changed, and (3) what drives the change.
RQ4--RQ5 examine external factors that may affect how journalists and scientists present certainty. We focus on (4) the prestige/quality of the journal, asking whether lower- or higher-quality journals differ in how certain their findings are, and (5) in the era of team science \cite{stokols2008science,hall2018science}, whether team size influences how certain the authors describe the results, given known differences in the nature of team science research outputs \cite{wu2019large,sud2016not,thelwall2016m}.

\myparagraph{Data and method} 
For RQ1 and RQ2, to control the effects of the content of the finding, we propose a method to match the same scientific finding in news and paper abstracts. For each extracted finding in a paper abstract, we identify the paraphrased findings in the corresponding news article reporting on that paper.  We first remove all punctuation and stop words and then stem all the words in each sentence. Next, we calculate the overlap and Jaccard similarity between each pair of findings in the news and abstract. We manually evaluated the matched findings and set word overlap $>=$3 and Jaccard similarity $>$0.3 as the threshold. Based on the findings from 52,406 news-paper pairs, we  identify 6,586 unique finding pairs from news and abstracts.  We manually annotated 70 matched finding pairs, and 63 (90\%) refer to the same science finding, indicating high precision of our matching process. Supplemental Material \Sref{app:matched-finding-sample} shows a random sample of the matched finding pairs. We construct separate regressions predicting the sentence-level (RQ1) and each aspect-level (RQ2) certainty in findings with the source of the finding (i.e., news or abstract). We further control the fields, author and affiliation ranking, journal impacts, finding length, and Flesch reading ease score. For RQ3, we construct a regression predicting the overall sentence-level certainty in news findings with the aspect-level certainty in the corresponding finding from the paper abstract. Except for all the IVs above, we further control the news outlet and the sentence-level certainty of the finding in the abstract. 

For RQ4 and RQ5, we construct a regression predicting the sentence-level certainty in 265,758 findings presented in 55,178 paper abstracts using journal impacts factors and the number of authors. Recognizing its limitations \citep{kurmis2003understanding}, we use journal impact factor as a proxy for the quality of science based on prior use \cite{saha2003impact}.
We include controls the field of the research, author, and affiliation ranking extracted from the Microsoft Academic Graph \cite{wang2019review}, finding length and Flesch reading ease score to remove the potential confounds. To test the connection between certainty in news findings and external factors, we construct another regression to predict the level of uncertainty in 72,013 findings presented in 27,000 news articles. Besides all the IVs regarding abstract and authors, we also control the outlet to remove potential confounds.

\myparagraph{RQ1: Are findings in news more certain than those in paper abstracts?} 
The regression analysis (details in Supplemental \tref{reg:source-sentence-level}) indicates that news descriptions have \textit{lower} overall sentence-level certainty than abstract descriptions of the same finding (p$<$0.01). Although existing studies suggest that science news tends to remove hedges and describe science findings with increased certainty \citep[e.g.,][]{weiss1988reporting,fahnestock2009rhetoric}, our study over the paired findings finds the opposite: findings in news are less certain compared with findings in abstract, even when controlling the content and many contextual factors. 

\begin{figure}[t]
 \includegraphics[width=2.9in]{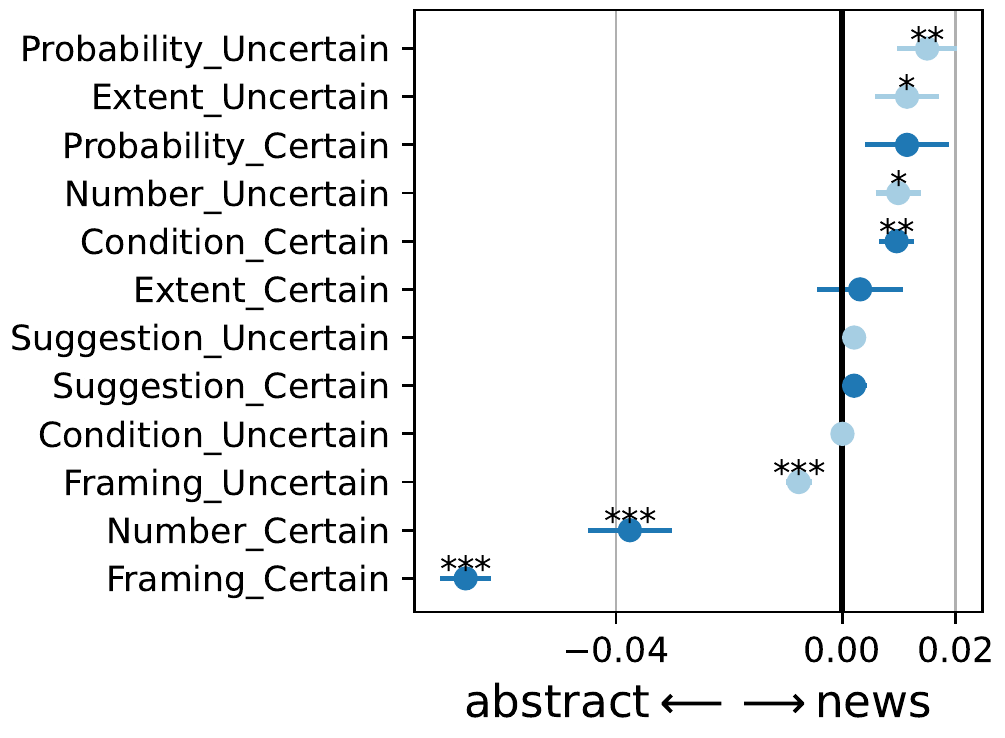}
  \caption{Controlling for multiple factors in RQ2 (e.g., topic, news outlet), the marginal effects show the relative probability of finding each aspect described in the abstract (left) versus news (right), revealing that some aspects like numeric certainty are much more likely to be described in one source.
  }
  \label{fig:aspect-certainty-source}
\end{figure}

\myparagraph{RQ2: Do journalists and scientists differ in their use of aspect-level certainty?} Yes, as shown in \fref{fig:aspect-certainty-source}, findings in abstracts are associated with more certainties about \textsc{Framing} and \textsc{Number}. Findings in news are associated with uncertainties about \textsc{Probability}, \textsc{Extent} and \textsc{Number}, indicating that the journalists tend to play down the certainty of some aspects, especially regarding numeric information. Existing studies suggest that laypeople with lower numeracy tend to focus more on narrative instead of numeric information \cite{dieckmann2009use}; one potential explanation for this difference is that journalists could be intentionally simplifying numerical information with hedges like ``roughly '' instead of the detailed number to better engage the lay audiences. Further, journalists are more likely to discard expressions of the scholar's uncertainty (\textsc{Framing}) when presenting the results, potentially aiming to make the work seem more objective. Our result suggests a potential mechanism for lower sentence-level certainty in news compared with abstracts. 

\begin{figure}[t]
 \includegraphics[width=2.7in]{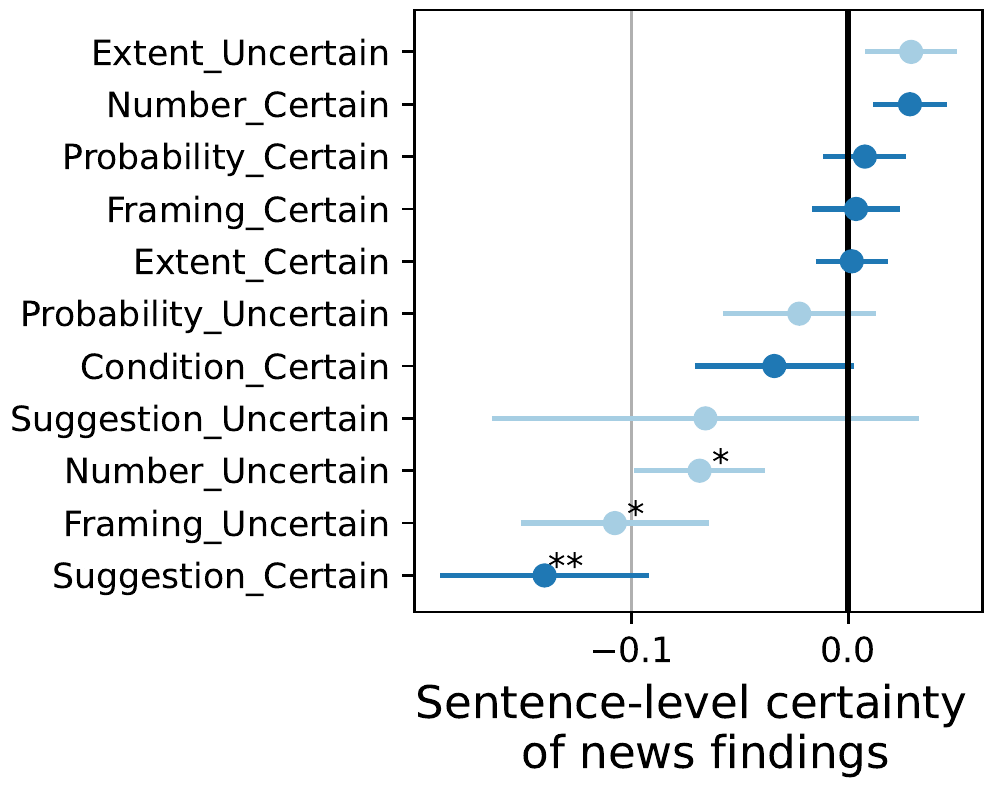}
  \caption{Averaged marginal effects of aspect-level certainty in abstract findings on the level of certainty for matched news findings. } 
  \label{fig:abs-types-news-levels}
\end{figure}

\myparagraph{RQ3: Does aspect-level certainty in abstract findings affect the sentence-level certainty in news findings?} As shown in figure \ref{fig:abs-types-news-levels}, uncertainty about \textsc{Number} and \textsc{Framing} are associated with decreased certainty in news findings, indicating that their uncertainty expressions are readily perceived by journalists. However, we also find that the certainty about \textsc{Suggestion} in abstracts are also associated with decreased certainty in news, suggesting that journalists may play down the certainty when presenting the findings involving suggestions or future actions even when it is certain. While existing studies suggest that journalists may  exaggerate the potential benefits of science \cite{wilson2010does}, our result indicates that journalists can be very careful when reporting findings involving suggestions or future actions.

\begin{figure}[t]
 \includegraphics[width=3.0in]{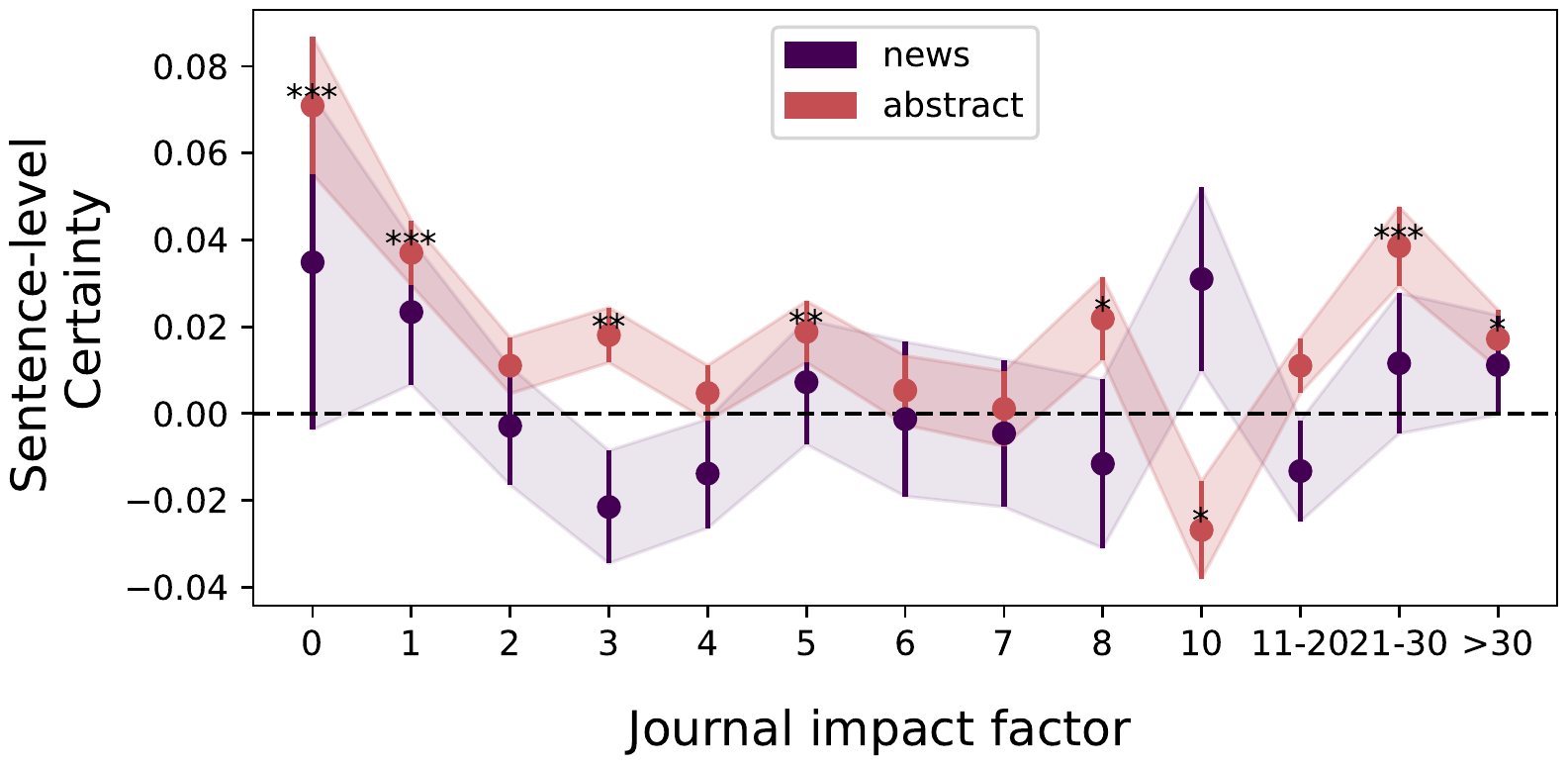}
 
  \caption{Averaged marginal effects of journal impact factors on level of certainty. Findings appearing on journals with lower impact factors are associated with higher levels of certainty. 
  }
  \label{fig:abs_news_journal_impact_level-of-uncertainty}
\end{figure}

\myparagraph{RQ4: Are findings in high-impact journals more certain than findings in low-impact journals?} No. As shown in Figure \ref{fig:abs_news_journal_impact_level-of-uncertainty}, findings in the \textit{lower}-impact journals are written with the highest level of certainty, while findings appearing in relatively higher-impact journals are described with comparatively less certainty. One potential explanation for this phenomenon is that high-quality papers published in journals with more strict reviewing processes\footnote{Journals with higher impact factors generally have longer reviews than low impact journals \citep[][p.~36]{publons2018}. } present certainty more precisely, which leads to a lower overall certainty compared with findings in low-impact journals. As a comparison, the certainty of findings written by journalists is not significantly associated with journal impact factors, suggesting that the prestige of a journal does not affect how journalists present scientific findings. %Existing studies 

\myparagraph{RQ5: Are findings from small teams more certain than findings from large teams?}
We find a linear relationship between the number of authors and the overall level of certainty in scientific findings (\fref{fig:abs_news_journal_impact_level-of-uncertainty}), even with controls for fields and authors.
Multiple mechanisms may explain this behavior. Larger teams may themselves be more capable of producing more certain results due to more individuals participating and checking results or due to the scale of the experiments capable in team science \cite{bozeman2004scientists,jeong2015collaborative}. Furthermore, our result also connects to \citet{wu2019large}'s finding that small teams generate new disruptive ideas while large teams tend to develop old, existing ideas, as new ideas are often associated with more uncertainties.
Alternatively, larger teams likely have more authors who are further away from the experiments and therefore may not be able to describe certainty with the same nuance and precision as smaller teams (or single individuals) who are intimately familiar with the methodological details \cite{mackenzie1998certainty}; this distance may increase the perception of certainty when writing. 
However, this linear trend does not persist in science news; \jpp{instead, the sentence-level certainty of findings in science news stays relatively steady across different numbers of authors. While team size has been found to be associated with the novelty and impact of science \cite{lee2015creativity,thelwall2016m, sud2016not}, our results indicate that the journalist is largely not influenced by the size of the research team in describing the certainty of their findings.}

Across these results, our study suggests that the journalists report scientific findings with \textit{less} certainty than scientists (RQ1). This result contradicts the existing findings that the journalists are overstating science \cite{weiss1988reporting,fahnestock2009rhetoric}. The fine-grained analysis over aspect-level certainty provides further details for such a change: journalists may play down the certainty about several core aspects of science findings like \textsc{Suggestion} even when they are certain in the abstract (RQ2,3). \jpp{Moreover, we find that the certainty of scientific findings in research articles varies with journal impact factors and team size, while such a pattern does not persist in science news (RQ4,5), suggesting that journalists may not alter scientific uncertainty according to these factors. }

\begin{figure}[t]

 \includegraphics[width=3.0in]{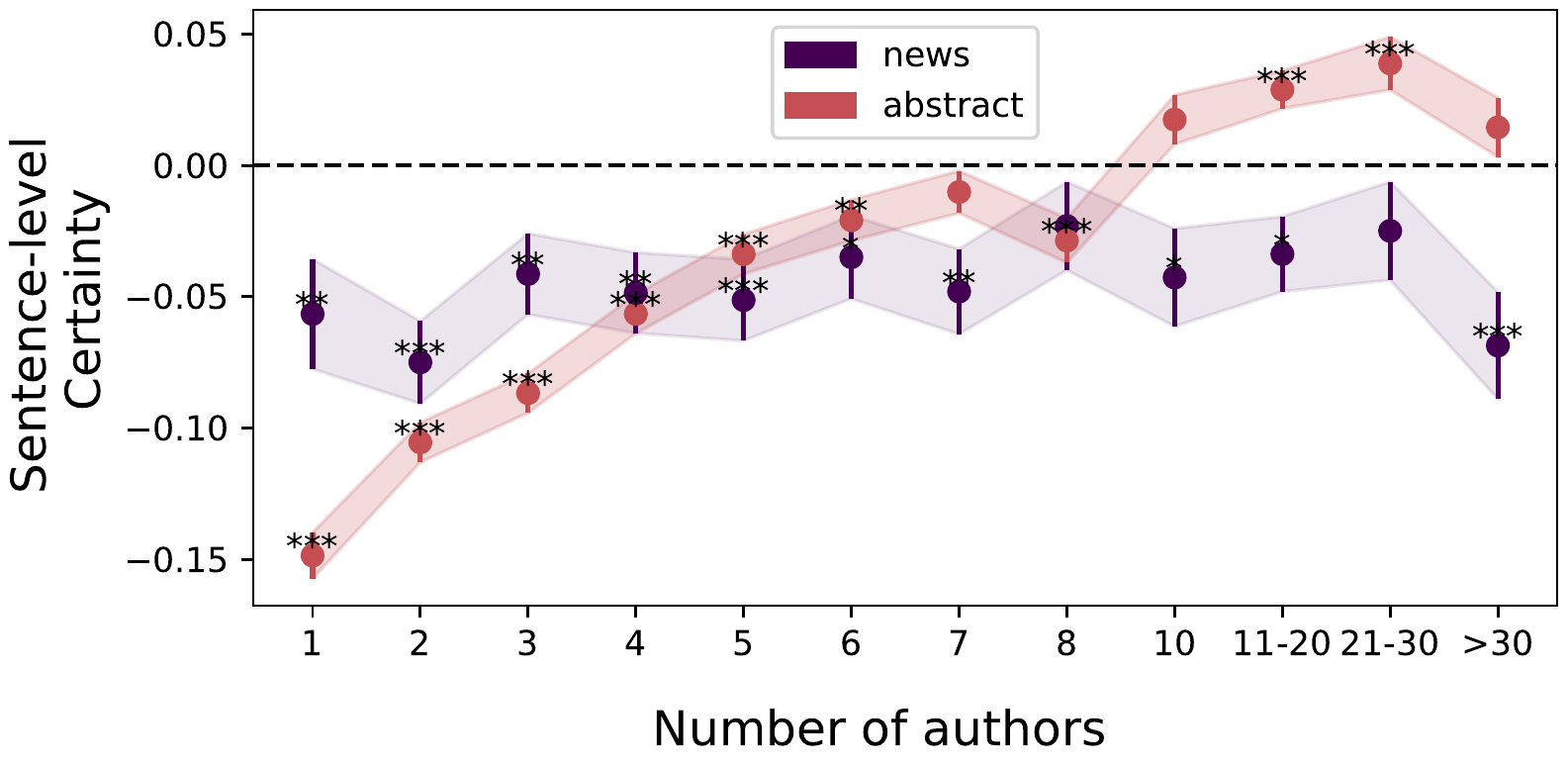}
  \caption{Averaged marginal effects of number of authors on level of certainty. Findings written by more authors are associated with a higher level of certainty, while such a pattern does not persist in science news.
  }
  \label{fig:abs_news_num_authors_level-of-uncertainty}
\end{figure}

\section{Discussion}
In this paper, we propose a new taxonomy, data, and models for certainty in science communications. Using the model, we analyzed a large scientific finding dataset and answered a series of important research questions on science communications. However, we also note the following limitations of our study. (1) We only use the abstract rather than the full text of research articles due to open access restrictions from copyright. Although authors normally present the core findings in the abstract, findings in abstracts could still be presented differently from findings in the main texts. (2) We use report verbs to extract findings in science news, which may miss findings that are presented without them. How to identify scientific findings in science news is still an open question and we call for future studies in this direction. (3) In our analysis, we use word-based heuristic methods (word overlap and Jaccard similarity) to match findings in news and abstracts, while the same scientific finding can be paraphrased with different sets of words. In future studies, we will develop better methods to identify paraphrases of scientific findings.

\section{Conclusion}

Our study represents a new step towards modeling certainty in text and demonstrates that sentence-level and aspect-level certainty are natural and feasible ways to model and annotate certainty. The proposed computational framework for certainty in scientific findings could support and inspire new studies on certainty in general language as well as new approaches to study science communication.

\section*{Acknowledgments}\label{sec:ack}
We thank Yumi Kim, Elizabeth Loeher, and Cassie Tian for data annotation. We thank Altmetric.com for sharing science journalism data and Hao Peng for building the preliminary data pipeline. We thank Misha Teplitskiy, Daniel Romero, Jian Zhu, Zuoyu Tian, Yian Yin for their helpful comments. We thank anonymous reviewers and area chairs for their helpful suggestions during the review process. 
This material is based upon work supported by the National Science Foundation under Grant No.~1850221, the Volkswagen Foundation, and Rackham Graduate Student Research Grant at the University of Michigan.

\bibliography{emnlp2020}
\bibliographystyle{acl_natbib}

\clearpage
\appendix

\section{Data Preprocessing}
\label{app:data-preprocessing}

\myparagraph{Altmetric mention data} Altmetric\footnote{https://www.altmetric.com/} tracks a variety of sources for mentions of research papers, including coverage from over 2,000 news outlets around the world. To control for differences in the frequency of scientific reporting and potential confounds from variations in journalistic practices across different countries, the list of news outlets was curated to 423 U.S.-based news media outlets, with each having at least 1,000 mentions in the Altmetric database. Location data for each outlet is provided by Altmetric. This initial dataset consists of 2.4M mentions of 521K papers by 1.7M news articles before 2019-10-06. Each mention in the Altmetric data has associated metadata that allows us to retrieve the original citing news story as well as the DOI for the paper itself.

\myparagraph{News processing} During data processing, we notice that some very long news articles are usually policy documents. Therefore, we removed news longer than 1392 words (top 5\%). To ensure that each news is specifically written about a single research paper's findings, we keep news only linked to one research paper. This leads to 128,942 news-paper pairs spanning 273 different news outlets and 57,807 different scientific articles. For all the news stories, we first remove references and paragraphs containing quotes as they might bias our analysis of uncertainty (e.g., a scientist describes their own work as uncertain in a quote).

\section{Model Details}
\label{sec:model_details}
We use scikit-learn version 0.23.1 to build the linear regression model \cite{pedregosa2011scikit}. Specifically, for the linear model, we use ridge regressor and classifier with default settings.  The built-in CountVectorizer of scikit-learn is used to vectorize the unigram, bigram, and trigram of each input question. The size of the bag-of-words feature vector is set as 40000.

For both the SciBERT and RoBERTa models, we use Hugging Face\footnote{\url{https://huggingface.co/}} transformers \cite{wolf2020transformers} and set the batch size as 128 and learning rate as 0.0001. We set \texttt{max\_len}=60. Adam \cite{kingma2014adam} is used for optimization. All the other hyperparameters and the model size are the same as the default \texttt{roberta-base} model and the \texttt{SciBERT} model. We train both models for 50 epochs and choose the model with the lowest loss on the validation set.  All the code, datasets, and parameters of our best-performing model are released and one could easily reproduce all the experiments.

\section{Additional Details on Extracting Scientific Findings}
\label{app:extracting-findings}
We use the following lexicons to extract scientific findings in news:

found that, find that, finds that, reveal that, reveals that, revealed that, suggest that, suggested that, suggests that, discover that, discovers that, discovered that, show that, shows that, showed that, conclude that, concludes that, concluded that, indicate that, indicates that, indicated that, claim that, claims that, claimed that, argue that, argues that, argued that

We manually annotate 50 extracted findings and only 1 of them does not fully counted as a scientific finding, indicating high precision of our approach. \tref{tab:extracted_finding_sample} presents the extracted findings from news and abstract. 

\section{Annotation agreement}
\label{app:annotation_agreemment}
\begin{figure}[t]
\includegraphics[width=3.0in]{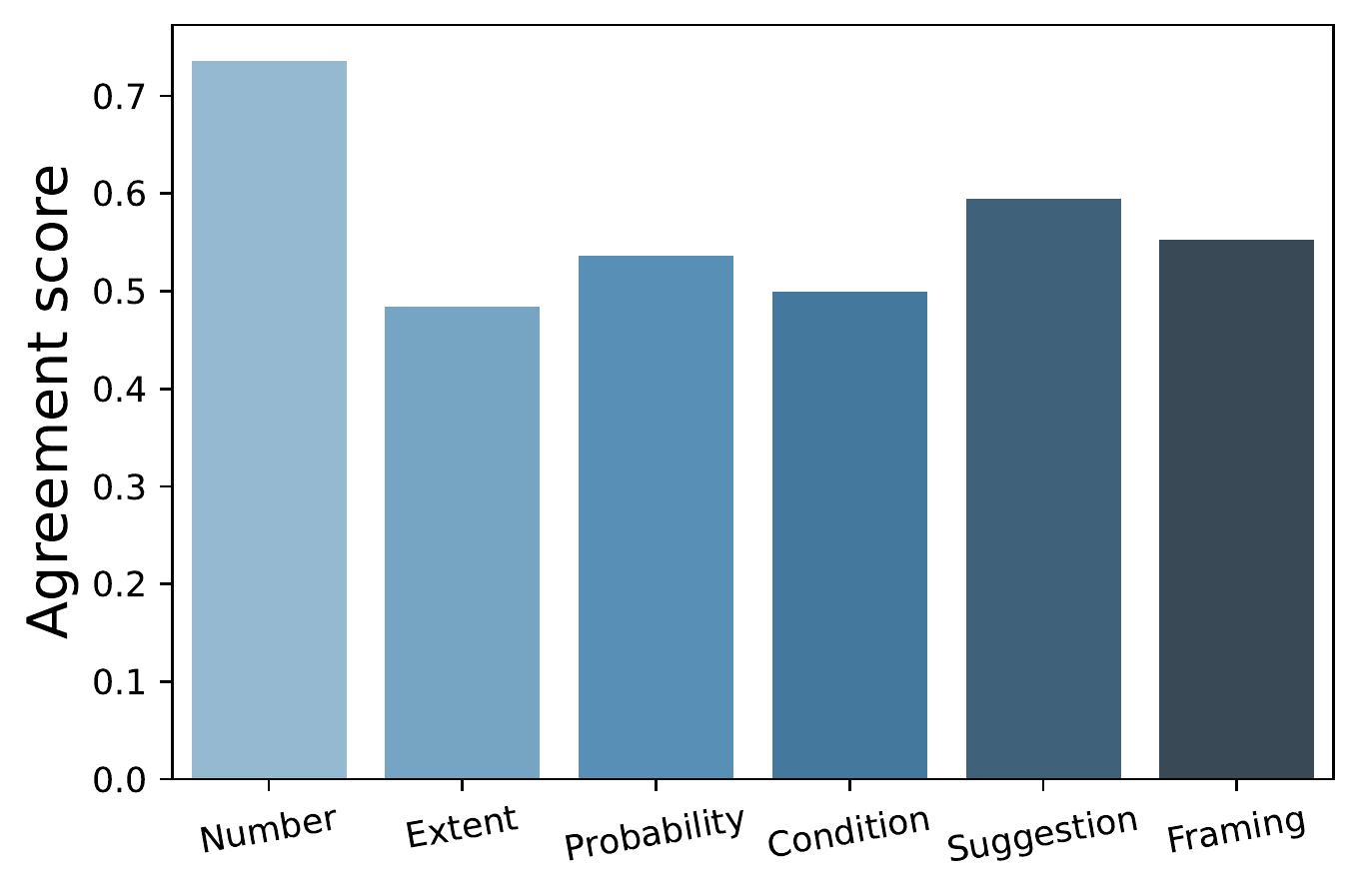}
\caption{Agreement scores for annotating aspect-level certainty}
\label{fig:agreement_type-of-uncertainty}
\end{figure}

\fref{fig:agreement_type-of-uncertainty} presents the Krippendorff's $\alpha$ for aspect-level certainty annotation. 

\section{Annotated data}
\label{app:annotated_data}
\begin{figure}[t]
%\vspace*{-0.7in}
\hspace*{-0.15in}
\centering
\includegraphics[width=2.7in]{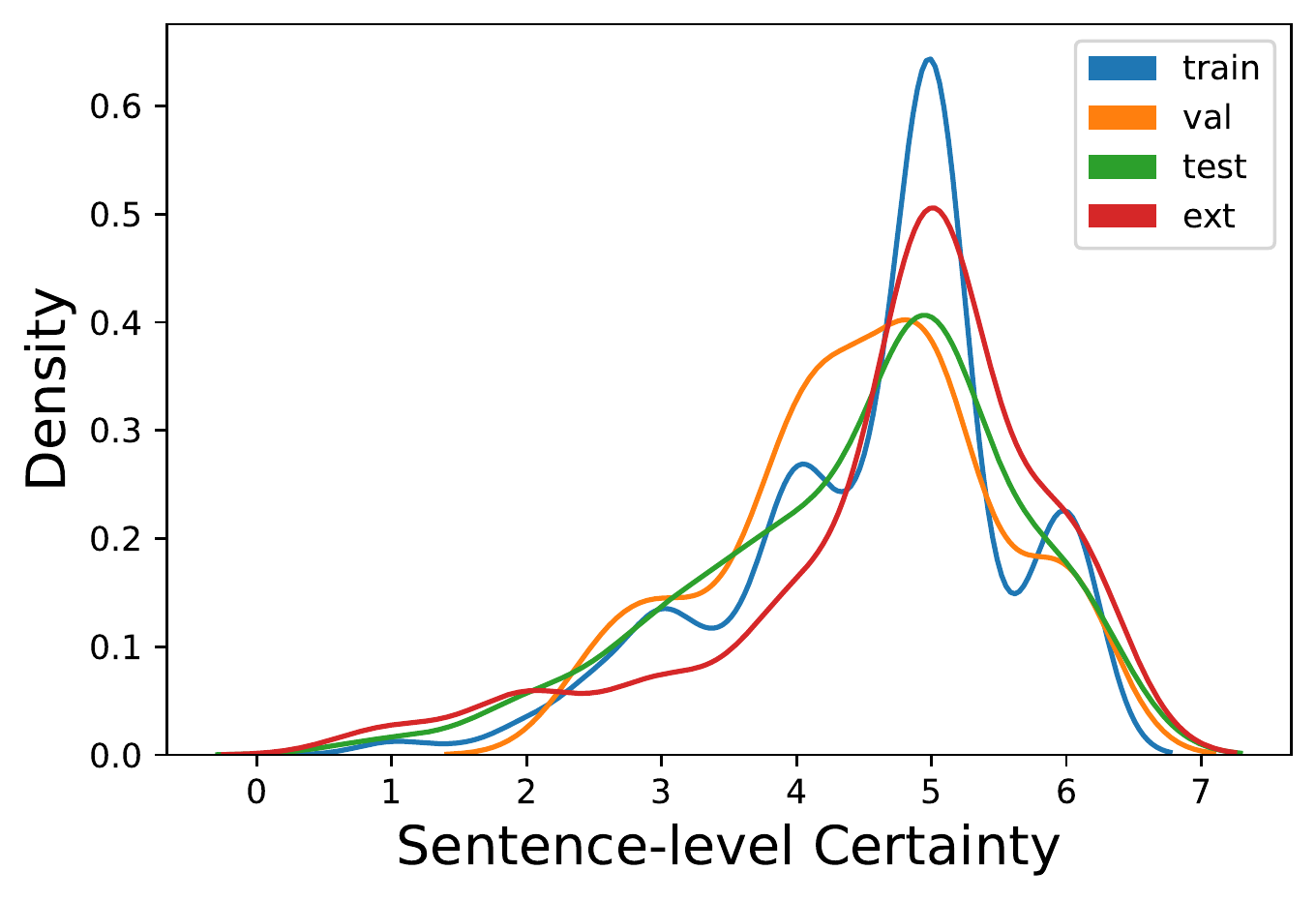}
%\vspace*{-0.10in}
 \caption{The distribution of sentence-level certainty score across the annotated dataset}
\label{fig:kdeplot_sentence-level-certainty}
\end{figure}

\begin{figure*}[t]
%\vspace*{-0.7in}
\hspace*{-0.15in}
\centering
\includegraphics[width=4.7in]{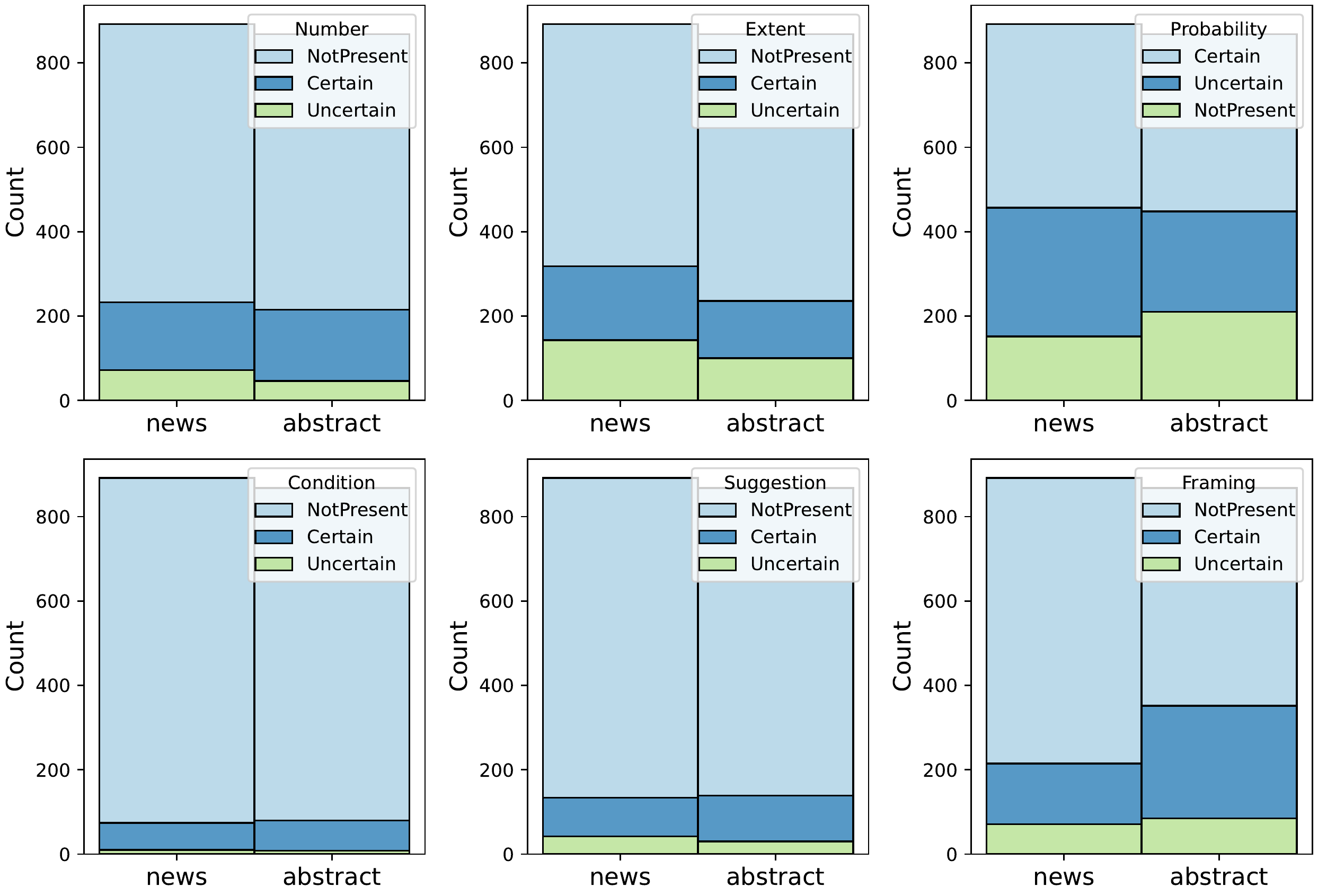}
%\vspace*{-0.10in}
 \caption{The distribution of aspect-level certainty scores in the annotated dataset}
\label{fig:aspects_distribution_source}
\end{figure*}

\fref{fig:aspects_distribution_source} presents the distribution of aspect-level certainty score in annotated dataset. \fref{fig:kdeplot_sentence-level-certainty} presents the distribution of sentence-level certainty across data splits.  \tref{tab:hedge_cnt_level_sample} and \tref{tab:aspect-certainty_sample} presents the annotated findings for sentence-level and aspect-level certainty.

\section{Matched finding samples}
\label{app:matched-finding-sample}
\tref{tab:matched_finding_sample} presents the samples for matched findings.

\section{Detailed regression results}
\label{app:all_reg_coefs}
\tref{reg:source-sentence-level} presents the regression results for RQ1. \tref{reg:aspect-certainty-source} presents the regression results for RQ2. \tref{reg:abs-types-news-levels} presents the regression results for RQ3. \tref{reg:if_team_size_news} and \tref{reg:if_team_size_abs} present the regression results for RQ4-5.

\begin{figure*}[t]
%\vspace*{-0.7in}
\hspace*{-0.15in}
\centering
\includegraphics[width=4.7in]{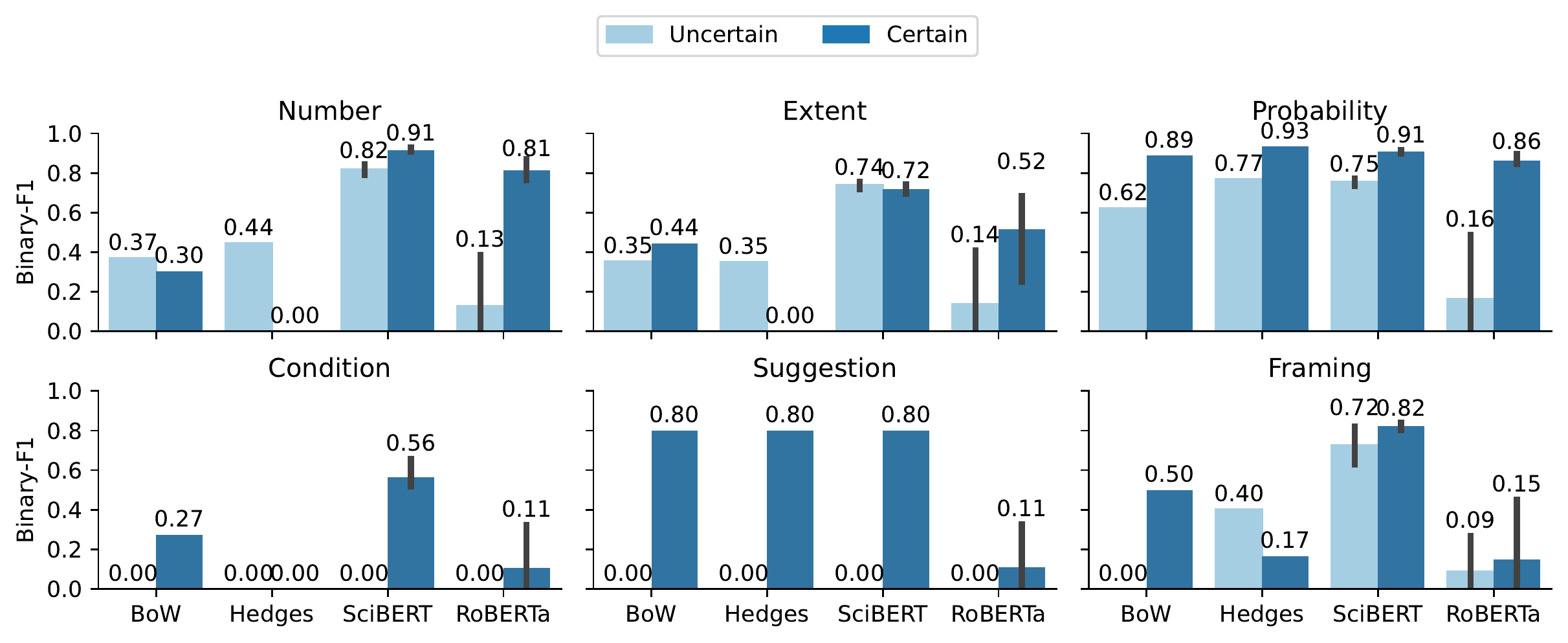}
%\vspace*{-0.10in}
 \caption{Aspect-level certainty prediction results over the random sample.}
\label{fig:model_perf_aspects_random_set}
\end{figure*}

\begin{table}[t]
\small

\newcommand{\tabincell}[2]{\begin{tabular}{@{}#1@{}}#2\end{tabular}}
\resizebox{0.45\textwidth}{!}{
\begin{tabular}{c|p{55mm}|c}
\toprule
\textbf{\# Hedges } & \textbf{Text} & \textbf{Score}\\
\midrule
4 & Based on these observations, we propose that the \textbf{apparent} receding contact angle \textbf{should} be used for characterizing superliquid-repellent surfaces \textbf{rather} than the \textbf{apparent} advancing contact angle and hysteresis. & 4.0 \\ 
%\hline
3 & The nondemented subjects with Alzheimer pathology \textbf{may} have had “preclinical” AD, or numerous cortical plaques \textbf{may} occur in some elderly subjects who \textbf{would} never develop clinical dementia. & 1.5 \\
%\hline
2 & These mouthwashes \textbf{may} be of modest benefit, it is \textbf{unclear} if a certain subset of patients showed a large response while others derived no benefit. & 2.0 \\
%\hline
1 & Circadian rhythm disruptions, such as jet lag, \textbf{might} be linked with an increased risk of cancer, she said. & 1.0 \\
%\hline
0 & Further research is necessary to understand whether this is a causal relationship. & 1.5 \\
\bottomrule
\end{tabular}
}
\caption{ Annotated sentence-level certainty for findings with different numbers of hedges. The number of hedges does not necessarily reflect the overall perception of certainty.}
\label{tab:hedge_cnt_level_sample}
\end{table}

\begin{table*}[th]
\small
\newcommand{\tabincell}[3]{\begin{tabular}{@{}#1@{}}#2\end{tabular}}
\centering
%\resizebox{\textwidth}{!}{
\begin{tabular}{p{60mm}|p{20mm}|p{20mm}|}
\toprule
\textbf{Finding} & \textbf{Uncertain} & \textbf{Certain} \\
\midrule
Arctic sea ice is declining at a rate of nearly 13 percent per decade. & 
\textsc{Number}, \textsc{Extent} & \textsc{Probability} \\ \hline
Some functions show decreases with potentially irreversible global impacts & \textsc{Extent}, \textsc{Probability} &  \\ \hline  
There were 365 cases of maternal sepsis, giving an incidence of severe maternal sepsis of 4.7 women per 100,000. & & \textsc{Number}, \textsc{Extent} \\ \hline
The practice may actually drive away qualified applicants who feel that their privacy has been compromised. & \textsc{Probability} &  \\ \hline
More research is needed on how cryptic post-copulatory and post-zygotic processes contribute to determining paternity and bridging the behavioural and genetic mating systems of viviparous species. & \textsc{Probability} & \textsc{Suggestion} \\ \hline

\bottomrule

\end{tabular}
%}
\caption{Annotated aspect-level certainty for sampled findings }
\label{tab:aspect-certainty_sample} 
\end{table*}

\begin{table*}[th]
\small

\newcommand{\tabincell}[3]{\begin{tabular}{@{}#1@{}}#2\end{tabular}}
\centering
%\resizebox{\textwidth}{!}{
\begin{tabular}{p{60mm}|p{60mm}|p{11mm}|p{10mm}}
\toprule
\textbf{Original text} & \textbf{Extracted finding} & \textbf{Keyword} & \textbf{Source} \\
\midrule
In addition, liver procurement data such as WIT showed that organs with less than 30 mins WIT led to significantly reduced yield, but no impact was found on viability. & Organs with less than 30 mins wit led to significantly reduced yield, but no impact was found on viability. & showed & news \\ \hline
I conclude that we live in one of infinitely many universes - one for each value of the gravitational constant. & We live in one of infinitely many universes - one for each value of the gravitational constant. & conclude & news \\ \hline
They found that the findings were specific to ADHD, with no associations observed between the other two disorders. & The findings were specific to adhd, with no associations observed between the other two disorders. & found & news \\ \hline
The Irish investigators of that meta-analysis found that methotrexate was associated with a small albeit statistically significant 10\% increase in the risk of all adverse respiratory events and an 11\% increase in the risk of respiratory infection. & Methotrexate was associated with a small albeit statistically significant 10\% increase in the risk of all adverse respiratory events and an 11\% increase in the risk of respiratory infection. & found & news \\
\hline

\bottomrule

\end{tabular}
%}
\caption{Extracted findings from news and abstracts}
\label{tab:extracted_finding_sample} 
\end{table*}

\begin{table*}[th]
\small

\newcommand{\tabincell}[3]{\begin{tabular}{@{}#1@{}}#2\end{tabular}}
\centering
%\resizebox{\textwidth}{!}{
\begin{tabular}{p{60mm}|p{60mm}|p{10mm}|p{10mm}}
\toprule
\textbf{News finding} & \textbf{Abstract finding} & \textbf{Jaccard Similarity} & \textbf{Word Overlap} \\
\midrule
For children with low self-esteem, high praise may be more harmful than helpful. & Inflated praise decreases challenge seeking in children with low self-esteem and has the opposite effect on children with high self-esteem. & 0.36 & 5 \\ \hline
Breast-feeding might be no more beneficial than bottle-feeding for 10 of 11 long-term health and well-being outcomes in children age 4 to 14. & Children aged 4 to 14 who were breast- as opposed to bottle-fed did significantly better on 10 of the 11 outcomes studied. & 0.35 & 7 \\ \hline
While our first impressions of educators might affect our ratings of them, ultimately the quality of their instruction matters the most in student evaluations. & Quality of instruction is the strongest determinant of student factual and conceptual learning, but that both instructional quality and first impressions affect evaluations of the instructor. & 0.39 & 7 \\ \hline
Even in the absence of symptoms, trauma may have an enduring effect on brain function. & Trauma has a measurable, enduring effect upon the functional dynamics of the brain, even in individuals who experience trauma but do not develop ptsd. & 0.38 & 6 \\ \hline
Being bullied may increase the risk for parasomnias. & Being bullied increases the risk for having parasomnias. & 0.67 & 4 \\ \hline
Dried fruits may lower the gi of white bread through displacement of high gi carbohydrate. & When displacing half the available carbohydrate in white bread, all dried fruit lowered the GI; however, only dried apricots $(GI\_=\_57\_\_\_5)$ showed a significant displacement effect $(P\_=\_0.025)$. & 0.36 & 8 \\
\bottomrule

\end{tabular}
%}
\caption{Matched findings from news and abstracts}
\label{tab:matched_finding_sample} 
\end{table*}

\onecolumn
 \footnotesize
 \setlength\LTleft{8mm}
\begin{longtable}{llrrrl}

%\resizebox{1.0\textwidth}{!}{
%\begin{tabular}{llrrrrrl}
\toprule
{} &                              index &         coeff &            SE & \multicolumn{2}{l}{p value} \\
\midrule
0   &                          Intercept &  4.611216e+00 &  7.331098e-02 &  0.000000e+00 &  *** \\
1   &                     source[T.news] & -3.152504e-02 &  1.030003e-02 &  2.212871e-03 &   ** \\
2   &                        finding\_len & -3.606883e-03 &  5.813032e-04 &  5.641478e-10 &  *** \\
3   &                flesch\_reading\_ease & -8.410694e-04 &  2.318612e-04 &  2.873043e-04 &  *** \\
4   &                  first\_author\_rank &  6.160764e-06 &  2.968990e-06 &  3.800262e-02 &    * \\
5   &                          affi\_rank &  4.904435e-06 &  2.961068e-06 &  9.768389e-02 &      \\
6   &                     journal\_impact &  1.831817e-03 &  4.162222e-04 &  1.085809e-05 &  *** \\
\end{longtable}
\onecolumn
 \footnotesize
 
 \setlength\LTleft{1mm}
\begin{longtable}{p{0.4cm}p{2.4cm}p{3.7cm}rrrp{0.2cm}}

\toprule
{} &                     DV &                                 IV &         coeff &            SE & \multicolumn{2}{l}{p value} \\
\midrule
0    &         Certain\_Number &                          Intercept & -3.857733e-01 &  5.288572e-02 &   3.171367e-13 &  *** \\
1    &         Certain\_Number &                     source[T.news] & -3.754525e-02 &  7.430325e-03 &   4.408625e-07 &  *** \\
2    &         Certain\_Number &                        finding\_len &  1.368212e-02 &  4.193456e-04 &  1.534606e-224 &  *** \\
3    &         Certain\_Number &                flesch\_reading\_ease &  4.748661e-03 &  1.672621e-04 &  4.235161e-172 &  *** \\
4    &         Certain\_Number &                  first\_author\_rank &  1.060523e-05 &  2.141796e-06 &   7.454170e-07 &  *** \\
5    &         Certain\_Number &                          affi\_rank & -1.366679e-06 &  2.136081e-06 &   5.223096e-01 &      \\
6    &         Certain\_Number &                        num\_authors &  1.048150e-03 &  2.912211e-04 &   3.204336e-04 &  *** \\
\end{longtable}
\onecolumn
 \footnotesize
 \setlength\LTleft{10mm}
\begin{longtable}{lp{5cm}rrrl}

%\resizebox{1.0\textwidth}{!}{
%\begin{tabular}{llrrrrrl}
\toprule
{} &                                          index &         coeff &            SE & \multicolumn{2}{l}{p value} \\
\midrule
0   &                                      Intercept &  1.964574e+00 &  1.730216e-01 &   1.361045e-29 &  *** \\
1   &                    abs\_Number\_STATE[T.Certain] &  2.844180e-02 &  1.696833e-02 &   9.375438e-02 &      \\
2   &                  abs\_Number\_STATE[T.Uncertain] & -6.859408e-02 &  3.020092e-02 &   2.316485e-02 &    * \\
3   &                    abs\_Extent\_STATE[T.Certain] &  1.616107e-03 &  1.655072e-02 &   9.222167e-01 &      \\
4   &                  abs\_Extent\_STATE[T.Uncertain] &  2.899808e-02 &  2.107222e-02 &   1.688306e-01 &      \\
5   &               abs\_Probability\_STATE[T.Certain] &  7.646274e-03 &  1.911305e-02 &   6.891295e-01 &      \\
6   &             abs\_Probability\_STATE[T.Uncertain] & -2.260107e-02 &  3.520943e-02 &   5.209589e-01 &      \\
\end{longtable}
\onecolumn
 \footnotesize
 \setlength\LTleft{8mm}
\begin{longtable}{llrrrl}

%\resizebox{1.0\textwidth}{!}{
%\begin{tabular}{llrrrrrl}

\toprule
{} &                                                 IV &         coeff &        SE & \multicolumn{2}{l}{p value} \\
\midrule
0   &                                        (Intercept) &  4.769169e+00 &  0.090361 &   0.000000e+00 &  *** \\
1   &                                        finding\_len & -8.585722e-03 &  0.000329 &  2.223725e-149 &  *** \\
2   &                                flesch\_reading\_ease & -1.233028e-03 &  0.000127 &   3.486418e-22 &  *** \\
3   &                                  first\_author\_rank &  6.810203e-06 &  0.000002 &   2.893930e-05 &  *** \\
4   &                                          affi\_rank & -1.955582e-07 &  0.000002 &   9.088450e-01 &      \\
5   &                             affi\_cateinternational &  6.176708e-03 &  0.006035 &   3.061173e-01 &      \\
6   &                                   affi\_cateunknown &  7.980791e-02 &  0.066196 &   2.279631e-01 &      \\
\end{longtable}
\onecolumn
 \footnotesize
 %\centering
 \setlength\LTleft{13mm}
\begin{longtable}{llrrrl}

%\resizebox{1.0\textwidth}{!}{
%\begin{tabular}{llrrrrrl}

\toprule
{} &                                 IV &     coeff &            SE & \multicolumn{2}{l}{p value} \\
\midrule
0   &                        (Intercept) &  4.702461 &  1.967952e-02 &   0.000000e+00 &  *** \\
1   &                        finding\_len & -0.005205 &  1.466497e-04 &  2.569752e-275 &  *** \\
2   &                flesch\_reading\_ease &  0.001294 &  5.116505e-05 &  6.128200e-141 &  *** \\
3   &                  first\_author\_rank &  0.000005 &  7.462000e-07 &   1.993841e-10 &  *** \\
4   &                          affi\_rank &  0.000004 &  8.059055e-07 &   1.423131e-06 &  *** \\
5   &             affi\_cateinternational &  0.002711 &  2.850670e-03 &   3.415584e-01 &      \\
6   &                   affi\_cateunknown & -0.059613 &  3.050521e-02 &   5.068047e-02 &      \\
\end{longtable}

\end{document}